\documentclass[11pt]{article}

\usepackage{color,soul}
\usepackage{amsmath,amsfonts,amssymb}
\usepackage{graphicx}
\usepackage{bbm}
\usepackage[ruled,linesnumbered,vlined]{algorithm2e}
\usepackage{booktabs}
\usepackage{multirow}
\usepackage[table]{xcolor}
\usepackage{comment}

\usepackage[]{acl}

\usepackage{times}
\usepackage{latexsym}

\usepackage[T1]{fontenc}

\usepackage[utf8]{inputenc}

\include{comment}
\usepackage{microtype}
\usepackage{todonotes}

\title{XLTime: A Cross-Lingual Knowledge Transfer Framework for Temporal Expression Extraction}

\author{Yuwei Cao\textsuperscript{1}, William Groves\textsuperscript{2}, Tanay Kumar Saha\textsuperscript{3}, Joel R. Tetreault\textsuperscript{2}, \\\textbf{Alex Jaimes\textsuperscript{2}, Hao Peng\textsuperscript{4}, Philip S. Yu\textsuperscript{1}} \\ \textsuperscript{1}Computer Science Department, University of Illinois Chicago \\ \textsuperscript{2}Dataminr Inc. \textsuperscript{3}Walmart Global Tech\\ \textsuperscript{4}BDBC, Beihang University\\ \texttt{\{ycao43,psyu\}@uic.edu}, \texttt{\{wgroves,jtetreault,ajaimes\}@dataminr.com} \\\texttt{tanaykumar.saha@walmart.com}, \texttt{penghao@buaa.edu.cn}}

\begin{document}
\maketitle
\begin{abstract}
Temporal Expression Extraction (TEE) is essential for understanding time in natural language. It has applications in Natural Language Processing (NLP) tasks such as question answering, information retrieval, and causal inference.  To date, work in this area has mostly focused on English as there is a scarcity of labeled data for other languages.  
We propose XLTime, a novel framework for multilingual TEE. XLTime works on top of pre-trained language models and leverages multi-task learning to prompt cross-language knowledge transfer both from English and within the non-English languages. XLTime alleviates problems caused by a shortage of data in the target language.
We apply XLTime with different language models and show that it outperforms the previous automatic SOTA methods on French, Spanish, Portuguese, and Basque, by large margins. XLTime also closes the gap considerably on the handcrafted HeidelTime method. 

\end{abstract}

\section{Introduction}


Temporal Expression Extraction (TEE) refers to the detection of \textit{temporal expressions} (such as dates, durations, etc., as shown in Table~\ref{table:sample_time_expressions}). It is an important NLP task \cite{uzzaman2013semeval} and has downstream applications in question answering \cite{choi2018quac}, information retrieval \cite{mitra2018introduction}, and causal inference \cite{feder2021causal}.
Most TEE methods work on English and are rule-based \cite{strotgen2013multilingual, zhong2017time}.
Deep learning-based methods \cite{ chen2019exploring, lange2020adversarial} are less common and report results on par with or inferior to the rule-based SOTAs.

Moreover, methods that work on other languages are rare, because of the scarcity of annotated data.  
We find that that there is considerable room for improving TEE, especially for low-resource languages. For example, the previous SOTA performance on the English TE3 dataset \cite{uzzaman2013semeval} is around $0.90$ in F1, while that on the Basque TEE benchmark \cite{altuna2016adapting} is merely $0.47$.  Recent deep learning methods, which have shown gains for many tasks, are underexplored for this important area of NLP. 
\begin{table}
\centering
   \begin{tabular}{l}
   \toprule
     In $\underbrace{\text{the last three months}}_{\textbf{Duration}}$, \text{net revenue rose 4.3\%}\\
     to \$525.8 million from \$504.2 million $\underbrace{\text{last year}}_{\textbf{Date}}$. \\
     The official news agency, which gives the $\underbrace{\text{daily}}_{\textbf{Set}}$ \\ tally of inspections, updated on $\underbrace{\text{Friday evening}}_{\textbf{Time}}$.\\
    \bottomrule
  \end{tabular}
 
\caption{Temporal expressions of different types (See Appendix~\ref{sec:appendix_types} for the definitions of the types).}
 \label{table:sample_time_expressions}
\vspace{-1.5em}
\end{table}

Developing an approach that can learn using the existing limited amount of training data is crucial for this field because of the effort required to develop high-quality rules for each language.
Thus we propose a cross-lingual knowledge transfer framework for multilingual TEE, namely, XLTime. 
We base our framework on pre-trained multilingual models \cite{devlin2018bert, conneau2019unsupervised}. 
We then use Multi-Task Learning (MTL) \cite{liu2019multi} to prompt knowledge transfer both from English and within the low-resource languages. For this, we design primary and secondary tasks. 
The primary task leverages the existing, annotated TEE data of the other languages. It transfers \textit{explicit knowledge} that tells the forms of the temporal expressions in a \textit{source language}. 
The secondary task maps the annotated source language TEE data samples to the target language using machine-translation tools, such as Google Translate, and acquires sentence-level labels (of the presence of one or more time expressions) from the original token-level labels.
It constructs training data in a weakly-supervised manner. The secondary task transfers \textit{implicit knowledge} by teaching the model to detect the presence of temporal expressions in text from the target language.
\noindent
\textbf{Contributions.}
\noindent \textbf{1)} We propose XLTime, which prompts cross-lingual knowledge transfer using MTL to address multilingual TEE. \textbf{2)} We show that XLTime outperforms the previous automatic SOTA methods by large margins on four languages including French (FR), Spanish (ES), Portuguese (PT), and Basque (EU), which are ``low-resource'' for the TEE task.
\textbf{3)} We show that XLTime also approaches the performance of the heavily handcrafted HeidelTime \cite{strotgen2013multilingual}, and XLTime even outperforms it on two languages (Portuguese and Basque). We make our code and data publicly available.\footnote{https://github.com/YuweiCao-UIC/XLTime}

\section{Related Work}

\begin{figure*}[t]
    \centering
    \includegraphics[width = 16cm]{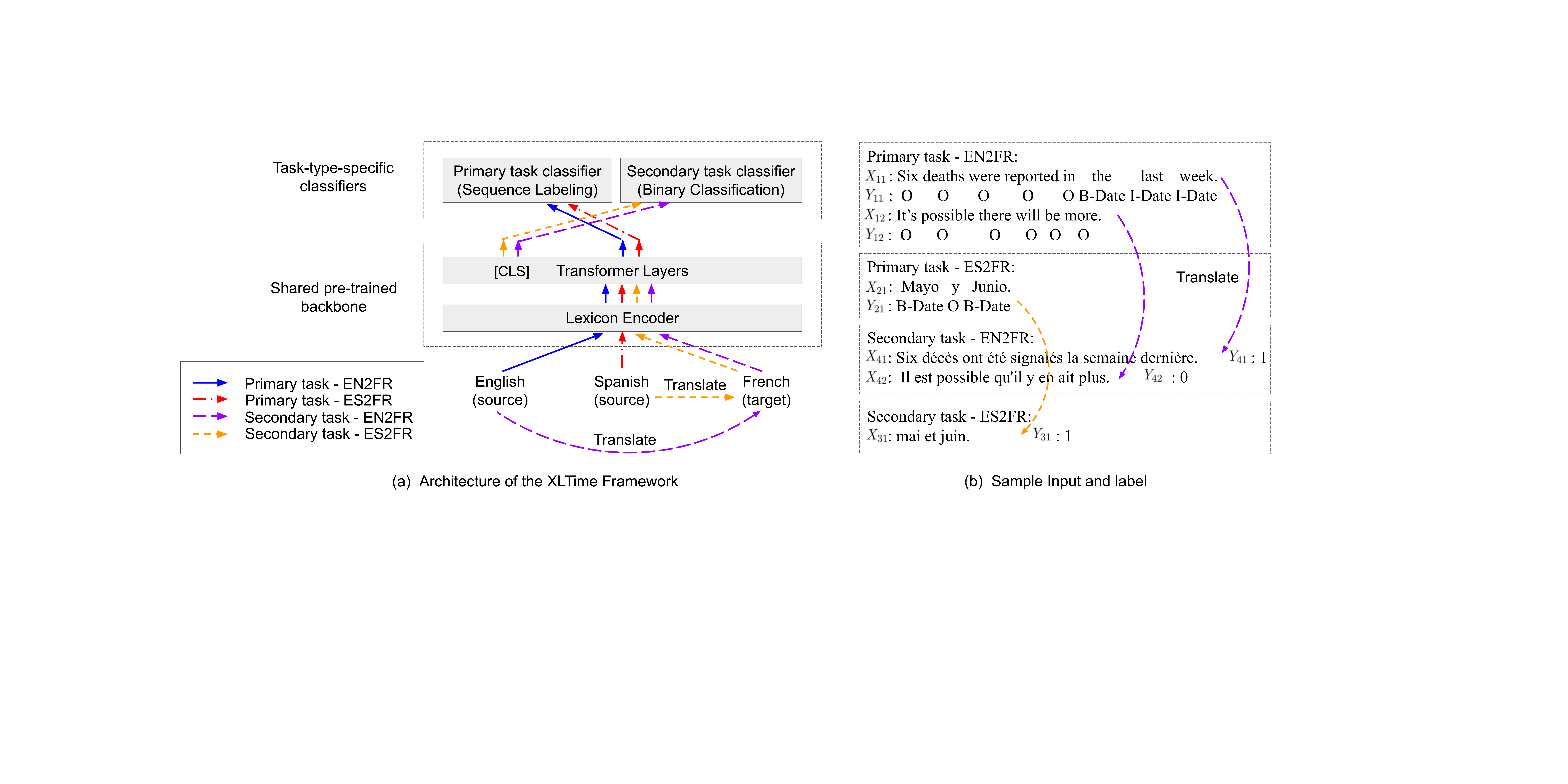}
    \caption{The architecture and sample training input of the proposed XLTime framework (\textit{best viewed in color)}. (a) shows how XLTime transfers knowledge from English (EN) and Spanish (ES) to French (FR) through the primary and the secondary tasks. (b) presents sample input of the tasks.}
    \label{fig:XLTime}
    \vspace{-1em}
\end{figure*}
While TEE is an important problem in NLP, there is relatively little work in the area, and most of this work focuses on English.  Prior art can be divided into two classes: rule/pattern-based and deep learning approaches.  In the first class, HeidelTime \cite{strotgen2013multilingual} 
is the top performing approach to date, and covers over a dozen languages.  It is driven by a collection of finely-tuned rules. The approach was later extended to more languages with HeidelTime-auto \cite{strotgen2015baseline}, which leverages language-independent processing and rules.  Other approaches include SynTime \cite{zhong2017time}, which is based on heuristic rules, and SUTIME \cite{chang2012sutime} and PTime \cite{ding2019pattern}, which leverages pattern learning.  

For the second class, \citet{laparra2018characters} proposes a model based on RNNs. \citet{chen2019exploring} uses BERT with a linear classifier.   \citet{lange2020adversarial} inputs mBERT embeddings to a BiLSTM with a CRF layer and outperforms HeidelTime-auto on four languages.  However, the reported performances of the deep learning-based methods are inferior to the rule-based ones, which is, in part, due to the complexity of the problem and training data paucity. In our work, we propose a new model which outperforms prior deep learning methods but also closes the gap considerably on HeidelTime, despite the data issues.

In addition, we are aware that applying label projection methods \cite{jain-etal-2019-entity} can be a straightforward way to address the data scarcity in non-English TEE. TMP \cite{jain-etal-2019-entity}, originally proposed for cross-lingual named entity recognition (NER) \cite{lample2016neural}, projects English data in IOB (Inside Outside Beginning) tagging format \cite{ramshaw1999text} to that of the other languages using machine translation, orthographic, and phonetic similarity packages. We show that the proposed XLTime, specifically designed to transfer temporal knowledge between languages, outperforms TMP by large margins.
\section{Proposed Method}

We formalize TEE as a sequence labeling task, similar to NER \cite{lample2016neural}. The architecture is shown in Figure~\ref{fig:XLTime}.


\subsection{Pre-trained Multilingual Backbone}
\label{model-backbone}
XLTime adopts SOTA multilingual models, i.e., mBERT \cite{devlin2018bert} and XLMR \cite{conneau2019unsupervised} as the backbone.
The pre-trained backbone contains lexicon and Transformer encoder layers as shown in Figure~\ref{fig:XLTime}(a). 
The backbone allows XLTime to acquire semantic and syntactic knowledge of various languages. The backbone is shared by the MTL tasks introduced in Section~\ref{model-tasks}.

\subsection{MTL-based Cross-Lingual Knowledge Transfer}
\label{model-tasks}
XLTime transfers knowledge from multiple \textit{source languages} to the low-resource \textit{target language}. The source languages include English and others for which TEE training data is available. 
We design \textit{primary} and \textit{secondary} tasks on top of the backbone to prompt \textit{explicit} and \textit{implicit} knowledge transfer. The primary task transfers knowledge that explicitly encodes the forms of the temporal expressions in a source language. It is formalized as sequence labeling and directly leverages the training data of the source language to train the backbone along with the primary task classifier, shown in Figure~\ref{fig:XLTime} (a). The primary task minimizes $\mathcal{L}_{sl}$:
\begin{equation}
\small
\mathcal{L}_{sl}=-\sum_{i=1}^{b}\sum_{j=1}^{m_i}\mathbbm{1}(y_{ij},c)log(softmax(\mathbf{W} \cdot \mathbf{x})),
\label{equation:primary}
\end{equation}
where $b$ is the total number of input sequences and $m_i$ is the length of the $i$th sequence. $\mathbf{x} \in \mathbb{R}^d$, output by the backbone, is the embedding of the $j$th token in the $i$th sequence. $d$ is its dimension. $c=argmax(\mathbf{W} \cdot \mathbf{x})$ and $y_{ij}$ are the predicted and ground-truth labels of the token. $\mathbf{W} \in \mathbb{R}^{|c|\times d}$ is the parameter of the primary task classifier. $|c|$ is the total number of unique ground-truth labels. $\mathbbm{1}(,)$ is 1 if its two arguments are equal and 0 otherwise. 


The secondary task implicitly reveals how the temporal expressions would be expressed in the target language. We translate the sequences in the source language training data into the target language using Google Translate
 (we observe similar results with AWS Translate).
%
The secondary task is formalized as binary classification,
where the input samples are the translated sequences and 
the labels are sentence-level indicators of whether or not the sequences contain temporal expressions (which can be easily inferred from the original labels).
This task tunes the model to learn the characteristics of temporal expressions in the target language in an implicit manner. 
It is weakly-supervised and requires no token-level labeling. 
It trains the backbone and the secondary task classifier by minimizing $\mathcal{L}_{bc}$:
\begin{equation}
\small
\mathcal{L}_{bc}=-\sum_{i=1}^{b}\mathbbm{1}(y'_{i},c')log(softmax(\mathbf{W'} \cdot \mathbf{x'})),
\label{equation:secondary}
\end{equation}
where $\mathbf{x}' \in \mathbb{R}^d$ is the sequence embedding output by the [CLS] of the backbone. $\mathbf{W}' \in \mathbb{R}^{2\times d}$ is the parameter matrix of the secondary task classifier. $c'=argmax(\mathbf{W'} \cdot \mathbf{x'})$ and $y'_{i}$ are the predicted and true sequence labels of the $i$th sequence.
We train XLTime concurrently on the primary and secondary tasks (further details found in Appendix~\ref{sec:appendix_training}).

\noindent 
\textbf{An Illustrative Example.} In Figure~\ref{fig:XLTime}, \textit{Primary task - EN2FR} and \textit{Secondary task - EN2FR} transfer knowledge from \textit{English} to \textit{French}. 
\textit{Primary task - EN2FR} reveals the exact forms of English temporal expressions using token-level labels ($Y_{11}$ and $Y_{12}$). \textit{Secondary task - EN2FR} takes the French translations ($X_{41}$ and $X_{42}$) of $X_{11}$ and $X_{12}$ as input. $Y_{41}$ and $Y_{42}$ indicate whether the sequences contain temporal expressions or not (can be inferred from $Y_{11}$ and $Y_{12}$). \textit{Secondary task - EN2FR} provides indirect knowledge about French temporal expressions. Similarly, \textit{Primary task - ES2FR} and \textit{Secondary task - ES2FR} transfer from \textit{Spanish} to \textit{French}.

\begin{table*}[th!]
\vspace{-1em}
\caption{\small The statistics of the datasets.}
  \centering
  \small
    \begin{tabular}{c|cccccccc}
   \toprule
     Lang & Dataset & Domain & \#Docs & \#Exprs & \#Dates & \#Times & \#Durations & \#Sets \\
    \midrule
     \centering{FR} & \citet{bittar2011french} & News & $108$ & $425$ & $227$ & $130$ & $52$ & $16$\\ 
     ES & \citet{uzzaman2013semeval} & News & $175$ & $1,094$ & $749$ & $57$ & $251$ & $37$\\ 
     PT & \citet{costa2012timebankpt} & News & $182$ & $1,227$ & $998$ & $41$ & $176$ & $12$\\
     EU & \citet{altuna2016adapting} & News & $91$ & $847$ & $662$ & $22$ & $151$ & $12$ \\
    \midrule
     & TE3 \cite{uzzaman2013semeval} & News & $276$ & $1,830$ & $1,471$ & $34$ & $291$ & $34$ \\ 
    EN& Wikiwars \cite{mazur2010wikiwars} & Narrative & $22$ & $2,634$ & $2,634$ & $0$ & $0$ & $0$\\ 
    & Tweets \cite{zhong2017time} & Utterance & $942$ & $1,128$ & $717$ & $173$ & $200$ & $38$ \\ 
    \bottomrule
  \end{tabular}
  \label{table:dataset_statistics_detail}
  \vspace{-1em}
\end{table*}

    

\begin{table}[t]

  \centering
  \small
    \begin{tabular}{l|cccc}
   \toprule
    
    Model & FR & ES & PT & EU \\
    \midrule
    \multicolumn{5}{l}{\texttt{\textbf{Automatic Baseline Models}}} \\
    HeidelTime-auto & 0.55 & 0.42 & 0.50 & 0.17 \\ 
    BiLSTM+CRF & 0.64 & 0.62 & 0.64 & 0.47 \\
    mBERT & 0.63 & 0.62 & 0.66 & 0.65 \\ 
    XLMR-base & 0.69 & 0.54 & 0.63 & 0.46 \\ 
    XLMR-large & 0.75 & 0.72 & 0.75 & 0.70 \\ 
    \midrule
    \multicolumn{5}{l}{\texttt{\textbf{Projection Method}}} \\
    TMP-mBERT & 0.56 & 0.23 & 0.66 & / \\ 
    TMP-XLMRbase & 0.55 & 0.23 & 0.64 & / \\ 
    TMP-XLMRlarge & 0.56 & 0.24 & 0.65 & / \\ 
    \midrule
    \multicolumn{5}{l}{\texttt{\textbf{Transfer from EN (Ours)}}} \\
    XLTime-mBERT & 0.73 & 0.71 & 0.67 & 0.76 \\ 
    XLTime-XLMRbase & 0.78 & 0.66 & 0.68 & 0.71 \\
    XLTime-XLMRlarge & 0.76 & 0.72 & 0.77 & \textit{0.78} \\
    \midrule
    \multicolumn{5}{l}{\texttt{\textbf{Transfer from EN and others (Ours)}}} \\
    XLTime-mBERT & 0.80 & \textbf{0.77} & \textit{0.80} & 0.77 \\ 
    XLTime-XLMRbase & \textit{0.82} & 0.72 & 0.73 & \textbf{0.79} \\
    XLTime-XLMRlarge & \textbf{0.84} & \textit{0.75} & \textbf{0.84} & \textbf{0.79} \\
    \bottomrule
    \multicolumn{5}{l}{\texttt{\textbf{Handcrafted Method}}}\\
    HeidelTime & 0.86 & 0.86 & 0.60 & / \\
    \bottomrule
  \end{tabular}
  \caption{\small Results for Multilingual TEE (Metric: F1).
}
  \vspace{-1em}
  \label{table:evaluation_low_resource_TEE_wo_type_f1}
\end{table}

\begin{table*}[t]

  \addtolength{\tabcolsep}{+3pt}
  \centering
  \small
    \begin{tabular}{l|c|ccc|c|ccc}
   \toprule
     Target Language & \multicolumn{4}{c|}{FR} & \multicolumn{4}{c}{ES} \\
     \hline
     Source Language(s)& EN & EN, EU & EN, PT& EN, ES & EN & EN, EU & EN, PT & EN, FR \\
    \hline
     XLTime-mBERT & 0.73 & 0.76 & 0.72 & \cellcolor{blue!25}\underline{0.80} & 0.71 & 0.72 & 0.72 & \cellcolor{blue!25}\underline{0.77} \\
     XLTime-XLMRbase & 0.78 & 0.76 & 0.78 & \cellcolor{blue!25}\underline{0.82} & 0.66 & 0.68 & \underline{0.71} & \cellcolor{blue!25}\underline{0.72} \\
     XLTime-XLMRlarge & 0.76 & \underline{0.81} & \underline{0.80} & \cellcolor{blue!25}\underline{0.84} & 0.72 & 0.72 & 0.75 & \cellcolor{blue!25}0.73 \\
    \midrule
    Target Language & \multicolumn{4}{c|}{PT} & \multicolumn{4}{c}{EU} \\
     \hline
     Source Language(s)& EN & EN, FR & EN, ES& EN, EU & EN & EN, PT & EN, ES & EN, FR \\
    \hline
     XLTime-mBERT & 0.67 & \cellcolor{blue!25}\underline{0.80} & \cellcolor{blue!25}0.70 & \underline{0.80} & 0.76 & 0.73 & 0.75 & 0.77 \\
     XLTime-XLMRbase & 0.68 & \cellcolor{blue!25}\underline{0.73} & \cellcolor{blue!25}0.63 & 0.56 & 0.71 & 0.74 & \underline{0.75} & \underline{0.79} \\
     XLTime-XLMRlarge & 0.77 & \cellcolor{blue!25}\underline{0.82} & \cellcolor{blue!25}\underline{0.84} & 0.74 & 0.78 & 0.79 & 0.79 & 0.77 \\
    \bottomrule
  \end{tabular}
  \caption{\small Low-resource language TEE with additional source languages (F1 scores
). The \colorbox{blue!25}{blue cells} are expected to, while the \underline{underlined cells} actually outperform (by $\geq$ 4\%) using EN as the only source language. }
  \vspace{-1em}
  \label{table:evaluation_source_languages_wot_f1}
\end{table*}

\section{Experiments}
This section evaluates the proposed XLTime framework. Section~\ref{sec:experimental_setup} introduces the datasets, models evaluated, metrics, and experimental settings. Section~\ref{sec:evaluation_results} quantitatively shows how XLTime alleviates data scarcity and prompts TEE performances. Section~\ref{sec:additional_languages} studies the effect of transferring knowledge from other languages in addition to English.  We also qualitatively show how XLTime transfers knowledge to the target languages in an error analysis in Appendix~\ref{sec:case_study}.
\subsection{Experimental Setup}
\label{sec:experimental_setup}
\noindent
\textbf{Datasets.} We use the English (EN), French (FR), Spanish (ES), Portuguese (PT), and Basque (EU) TEE benchmark datasets. Table~\ref{table:dataset_statistics_detail} shows dataset statistics. 
For each target language, we split its dataset with 10\% for validation and 90\% for test.
For each source language (applicable to XLTime), we use the whole dataset for training. 
\noindent
\noindent 
\textbf{Baselines.} 
We evaluate against rule-based, deep learning-based, and entity projection-based methods.  
We compare to the handcrafted HeidelTime \cite{strotgen2013multilingual} and its automatically extended version, HeidelTime-auto \cite{strotgen2015baseline}.  We also compare to deep learning methods: BiLSTM+CRF \cite{lange2020adversarial}, mBERT, base and large versions of XLMR. 
In addition, we compare to TMP \cite{jain-etal-2019-entity}, a cross-lingual label projection method which relies on machine translation as well as orthographic and phonetic similarity packages (unavailable for EU). 



\noindent \textbf{Our Approaches.}  We test several variants of our proposed model, which can be broken into two classes:  1) Cross-lingual transfer from EN. We apply XLTime on mBERT, base and large versions of XLMR and use EN as the only source language.
2) Cross-lingual transfer from EN and others. We transfer from other languages in addition to EN. 

\noindent
\textbf{Evaluation Metrics.}
We report F1
in \textit{strict match} \cite{uzzaman2013semeval}, i.e., all its tokens must be correctly recognized for an expression to be counted as correctly extracted.

We follow the setting in prior work of evaluating ``without type'' 
and report the results without considering the types of the temporal expressions (e.g., for `see you tomorrow', a prediction such as `O O B-Duration' would be counted as correct, though the proper labeling would be `O O B-Date').\footnote{
We do note that the temporal expression field should ultimately evaluate on the more complex task of identifying temporal expressions as well as their types.
This is in the spirit of the annotations and is in line with other sequence labeling tasks, such as NER. Therefore, we also experiment with the ``with type'' setting and show results in Appendix~\ref{sec:with_type}.  In both settings, the observations made in Sections~\ref{sec:evaluation_results} and \ref{sec:additional_languages} hold and XLTime outperforms the previous SOTAs by large margins.  
}

\noindent
\textbf{Experimental Setting.} We set $d$, the embedding dimension, to be consistent with the pre-trained multilingual backbone's dimension (768 for the base version language models and 1024 for large versions).
We use AdamW \cite{loshchilov2017decoupled} with a learning rate of $7e^{-6}$ and warm-up proportion of $0.1$. We train the models for 50 epochs and use the best model as indicated by the validation set for prediction. All datasets are transformed into IOB2 format to fit the sequence labeling setting.
All the deep learning methods are trained on English TEE datasets, validated and evaluated on low-resource languages.
For BiLSTM+CRF, we use the hyperparameters as suggested in the original paper \cite{lange2020adversarial}.
For TMP, we use it to project the English dataset to the target languages, take the projected data to train the language models, then validate and evaluate on the target languages. 
We perform a grid search over $\{$0.05, 0.1, 0.15, 0.25, 0.5$\}$ to tune $\delta$, the similarity score threshold of TMP,
and present the best performance.
We repeat all experiments for 5 times and report the mean result.
All experiments are conducted on a 64 core Intel Xeon CPU E5-2680 v4@2.40GHz
with 512GB RAM and 1×NVIDIA Tesla P100-PICE GPU.

\subsection{Multilingual TEE}
\label{sec:evaluation_results}
We evaluate XLTime on multilingual TEE (see Table~\ref{table:evaluation_low_resource_TEE_wo_type_f1} and Appendix~\ref{sec:full_table_low_TEE}). We observe:
\textbf{1)} XLTime-XLMRlarge outperforms the strongest automatic baseline by up to 
9\% in F1 on all languages.
It even outperforms the handcrafted HeidelTime method by a sizable margin (24\% in F1) in PT.
\textbf{2)} Applying XLTime improves upon the vanilla language models, even when transferring knowledge only from EN. E.g., XLTime-XLMRbase outperforms XLMR-base by 13\%, 22\%, 8\%, and 54\% in F1 on FR, ES, PT, and EU.
\textbf{3)} Introducing additional source languages to XLTime further improves the performance:
the F1 improves by up to 19\%, 11\%, and 11\% for XLTime-mBERT, XLTime-XLMRbase, and XLTime-XLMRlarge.
\textbf{4)} HeidelTime is a very hard baseline to beat given the time and care that went into developing language-specific rules.  
However, XLTime approaches its performance for FR and ES, outperforms it for PT, and makes 
predictions for EU (where HeidelTime has no rules). Note the previous automatic SOTA, XLMR-large, also outperforms HeidelTime for PT, but not as significantly. This shows that the automatic methods are increasingly promising for the non-English TEE task. \textbf{5)} XLTime-XLMRlarge improves upon XLMR-large by a large margin (11\% in F1) in EU. For FR, ES, and PT, the improvements are smaller. This may because XLMR-large, compared to mBERT and XLMR-base, is already very knowledgeable (especially in FR, ES, and PT, which are more common than EU). Therefore, applying XLTime may not provide much improvement (in contrast, applying XLTime on mBERT and XLMR-base dramatically boosts F1 by 8-54\%).
\textbf{6)} TMP performs poorly probably because the falsely projected entities can mislead the language models. Specifically, the token-by-token machine translation and matching process of TMP does not work well for temporal entities, especially when the target language TEs contain definite articles, prepositions, etc., that do not have explicit matches in the source language. E.g., EN TE `yesterday morning' can be correctly map to FR TE `hier matin' ('yesterday' to `hier' and `morning' to `Matin') but not to EU TE `ayer por la mañana' ('yesterday' to `ayer' and `morning' to `Mañana', leaving `por' and `la' unmatched).

\subsection{Transfer Knowledge from Additional Languages}
\label{sec:additional_languages}
We also study the effect of transferring additional knowledge from a low-resource language in addition to English, see Table~\ref{table:evaluation_source_languages_wot_f1} and 
Appendix~\ref{sec:full_table_low_TEE}. Our assumption is that similar languages (FR, ES, and PT) would help each other (one exception is PT, as the published dataset is EN text translated to PT and we, therefore, don't expect machine translation to provide additional knowledge). We observe: \textbf{1)} In most cases, transferring additional knowledge from similar languages (blue cells) does dramatically improve performance (underlined cells), with F1 increasing by up to 13\%. \textbf{2)} In some rare cases, negative knowlege transfer \cite{wu2020understanding}
occurs as adding source languages hurts performance (e.g., EN, ES $\rightarrow$ PT scores lower than EN $\rightarrow$ PT for XLTime-XLMRbase). We hypothesize this is related to the quality of the datasets and plan to address this in the future. 

\section{Conclusion}
\label{Conclusion}
We propose XLTime for multilingual language TEE in low-resource scenarios. It is based on language models and leverages MTL to prompt cross-language knowledge transfer. It greatly alleviates the problems caused by the shortage in training data and shows results superior to the previous automatic SOTA methods on four languages. It also approaches the performance of a highly engineered rule-based system.
\section*{Acknowledgements}
This work is supported in part by NSF under grants III-1763325, III-1909323, III-2106758, SaTC-1930941 and the S\&T Program of Hebei through grant 21340301D. For any correspondence, please refer to Hao Peng.

\bibliography{anthology,custom}
\bibliographystyle{acl_natbib}



\clearpage
\newpage

\newpage
\appendix


\section{Types of the Temporal Expressions}
\label{sec:appendix_types}
According to ISO-TimeML \cite{pustejovsky2010iso}, the TEE dataset annotation guideline, there are four types of temporal expressions, i.e., \textit{Date}, \textit{Time}, \textit{Duration}, and \textit{Set}. 
\textit{Date} refers to a calendar date, generally of a day or a larger temporal unit; \textit{Time} refers to a time of the day and the granularity of which is smaller than a day; \textit{Duration} refers to the expressions that explicitly describe some period of time; \textit{Set} refers to a set of regularly recurring times \cite{pustejovsky2010iso}.

\begin{algorithm}[!h]
\SetAlgoVlined
\caption{Training XLTime }
\label{alg:mini-batch training}
// Initialize model.

Load the parameters
from a pre-trained multilingual model. 

Initialize $\mathbf{W}$ and $\mathbf{W'}$ randomly.

// Prepare task data.

\For {t in $\{primary, secondary\}$} {
    Split the data of task $t$ into mini-batches $B_t$
}
$B$ = $B_{primary} \cup B_{secondary}$

\For {e in 1, ..., epoch}{
    Randomly shuffle $B$
    
    // $b_t$ is a mini-batch of task $t$
    
    \For {$b_t$ in $B$}{
        \eIf{$t$ is a primary task}{
            $\mathcal{L}_{sl} = $ Equation~\ref{equation:primary}
        }{
            $\mathcal{L}_{bc} = $ Equation~\ref{equation:secondary}
        }
        Compute gradient and update model parameters
    }  
}

\end{algorithm}

\section{The Training Procedure}
\label{sec:appendix_training}
We adopt mini-batch-based stochastic gradient descent (SGD) to train XLTime, as shown in Algorithm~\ref{alg:mini-batch training}. To concurrently train on the primary and secondary tasks, we split the training data of both tasks into mini-batches and randomly take one mini-batch at each step. We then calculate loss using that mini-batch and update the parameters of the shared backbone as well as the task-type-specific classifier. The classifier of the other task type is unaffected.



\section{Full Results for Low-resource Language TEE}\label{sec:with_type}
Table~\ref{table:evaluation_low_resource_TEE_wt} shows the full results for low-resource language TEE with/without considering the types of the temporal expressions. Note that the superiority of our proposed XLTime over the previous automatic SOTA still holds.

\section{Full Results for Low-resource Language TEE with Additional Source Languages}
\label{sec:full_table_low_TEE}
Tables~\ref{table:evaluation_source_languages_wt_f1_pr_rc}~and~\ref{table:evaluation_source_languages_wot_pr_rc} show the full results for low-resource language TEE with additional source languages. 

\section{Comparative Error Analysis} 
\label{sec:case_study}
\begin{table*}[!h]
\caption{\small mBERT vs. XLTime-mBERT (transfer from EN) frequent (count $\geq$ 10) errors.} 
  \centering
  \small
    \begin{tabular}{c|ccccc}
   \toprule
     Error Desc. & FR TEs & EN translations & mBERT results (wrong)	& XLTime results (correct) & counts \\
    \midrule
     
     \multirow{2}{22mm}{fail to recognize `hier (yesterday)'} 
     & 
     \multicolumn{1}{l}{hier soir} & \multicolumn{1}{l}{last night} & \multicolumn{1}{l}{O B-TIME} & \multicolumn{1}{l}{B-TIME I-TIME} & \multirow{2}{*}{30}\\
     &
     \multicolumn{1}{l}{hier} & \multicolumn{1}{l}{yesterday} & \multicolumn{1}{l}{O} & \multicolumn{1}{l}{B-DATE} \\
     
     \midrule
     
     \multirow{3}{22mm}{fail to recognize vague time span} 
     & 
     \multicolumn{1}{l}{désormais} & \multicolumn{1}{l}{from now on} & \multicolumn{1}{l}{O} & \multicolumn{1}{l}{B-DATE} & \multirow{3}{*}{6}\\
     &
     \multicolumn{1}{l}{longtemps} & \multicolumn{1}{l}{long time} & \multicolumn{1}{l}{O} & \multicolumn{1}{l}{B-DURATION} \\
     &
     \multicolumn{1}{l}{toute l'année} & \multicolumn{1}{l}{all year} & \multicolumn{1}{l}{O O} & \multicolumn{1}{l}{B-SET I-SET} \\
     
     \midrule
     
     \multirow{3}{22mm}{fail to recognize definite articles and adjectives} 
     & 
     \multicolumn{1}{l}{le 3 août} & \multicolumn{1}{l}{August 3} & \multicolumn{1}{l}{O B-DATE I-DATE} & \multicolumn{1}{l}{B-DATE I-DATE I-DATE} & \multirow{3}{*}{10}\\
     &
     \multicolumn{1}{l}{la nuit} & \multicolumn{1}{l}{the night} & \multicolumn{1}{l}{O O} & \multicolumn{1}{l}{B-TIME I-TIME}\\
     &
     \multicolumn{1}{l}{lundi prochain} & \multicolumn{1}{l}{next Monday} & \multicolumn{1}{l}{B-DATE O} & \multicolumn{1}{l}{B-DATE I-DATE}\\
     
    \bottomrule
  \end{tabular}
  \label{table:mBERT_vs_XLTime_EN}
\end{table*}

\begin{table*}[t]
\caption{\small XLTime-mBERT (transfer from EN) vs.  XLTime-mBERT (transfer from EN and ES) frequent (count $\geq$ 8) errors.} 
  \centering
  \small
    \begin{tabular}{c|ccccc}
   \toprule
     Error Desc.  & FR TEs & EN translations & EN results (wrong)	& EN and ES results (correct) & counts \\
    \midrule
     
     \multirow{3}{22mm}{fail to recognize definite articles and prepositions} 
     & 
     \multicolumn{1}{l}{en été} & \multicolumn{1}{l}{in summer} & \multicolumn{1}{l}{B-DATE I-DATE} & \multicolumn{1}{l}{O B-DATE} & \multirow{3}{*}{20}\\
     &
     \multicolumn{1}{l}{le 13 février} & \multicolumn{1}{l}{February 13} & \multicolumn{1}{l}{O B-DATE I-DATE} & \multicolumn{1}{l}{B-DATE I-DATE I-DATE} \\
     &
     \multicolumn{1}{l}{de dimanche} & \multicolumn{1}{l}{of Sunday} & \multicolumn{1}{l}{B-DATE I-DATE} & \multicolumn{1}{l}{O B-DATE} \\
     
     \midrule
     
     \multirow{2}{22mm}{wrong token types} 
     & 
     \multicolumn{1}{l}{mardi} & \multicolumn{1}{l}{Tuesday} & \multicolumn{1}{l}{B-TIME} & \multicolumn{1}{l}{B-DATE} & \multirow{2}{*}{18}\\
     &
     \multicolumn{1}{l}{quelques jours} & \multicolumn{1}{l}{A few days} & \multicolumn{1}{l}{B-DATE I-DATE} & \multicolumn{1}{l}{B-DURATION I-DURATION} \\
     
     \midrule
     
     \multirow{2}{22mm}{recognized extra TEs} 
     & 
     \multicolumn{1}{l}{quotidiens} & \multicolumn{1}{l}{daily} & \multicolumn{1}{l}{B-SET} & \multicolumn{1}{l}{O} & \multirow{2}{*}{8}\\
     &
     \multicolumn{1}{l}{la saison} & \multicolumn{1}{l}{the season} & \multicolumn{1}{l}{B-DATE I-DATE} & \multicolumn{1}{l}{O O}\\
     
    \bottomrule
  \end{tabular}
  \label{table:XLTime_EN_vs_XLTime_EN_ES}
\end{table*}

This section qualitatively shows how the proposed XLTime framework transfers knowledge to the target language. Specifically, we show how the errors made by the vanilla multilingual models can be fixed by applying XLTime. We also show how applying XLTime on other languages in addition to English would help fix more errors. 

We compare mBERT and XLTime-mBERT (transfer from EN) on FR TEE. 
Table~\ref{table:mBERT_vs_XLTime_EN} summarizes cases where mBERT fails while XLTime-mBERT gives correct predictions. 
We can tell that 
XLTime-mBERT learns `hier (yesterday)', which is not understood by the mBERT model. XLTime-mBERT also learns to recognize vague time spans such as `désormais (from now on)' and `longtemps (long time)', which are missed by the mBERT model. Moreover, compared to mBERT, XLTime-mBERT understands FR grammar better, as it recognizes the roles of definite articles and adjectives, such as `le (the)' and `prochain (next)', in TEs. In a word, the proposed 
XLTime framework helps connect the concepts in EN to the corresponding ones in FR.

To show how applying XLTime on extra source languages would help fix more errors, we compare XLTime-mBERT (transfer from EN) and XLTime-mBERT (transfer from EN and ES) on FR TEE. Table~\ref{table:XLTime_EN_vs_XLTime_EN_ES} summarizes the TEs that the former fails while the latter gives correct predictions. We can tell that by leveraging ES as an additional source language, XLTime-mBERT better masters FR grammar. Specifically, it learns to recognize definite articles and prepositions that share similar (e.g., `le/los') or identical (e.g., `de' and `en') forms in ES and FR. It can also better distinguish TEs of different types (e.g., it learns that `quelques jours (a few days)' is a \textit{Duration}, instead of a \textit{Date}). One interesting fact is, when transferring solely from EN, the model recognizes some extra TEs that are not in the ground truth of the FR dataset. This is because of an inconsistency in data labeling: `daily' is considered as a \textit{Set} in the EN dataset, while its counterpart, `quotidiens' is overlooked in the FR dataset. The proposed XLTime framework eliminates the needs of manually labeling multiple datasets and therefore, can be applied to minimize data label inconsistency.
\addtolength{\tabcolsep}{+2pt}
\begin{table*}[t]
\caption{\small Multilingual TEE results (w/ type | w/o type).}
  \centering
  \small
    \begin{tabular}{l|ccc|ccc|ccc|ccc}
   \toprule
    
    \multirow{2}{*}{w/ type} &  \multicolumn{3}{c}{FR} & \multicolumn{3}{c}{ES} & \multicolumn{3}{c}{PT} & \multicolumn{3}{c}{EU}\\
    \cline{2-4} \cline{5-7} \cline{8-10} \cline{11-13}
    \multirow{2}{*}{} & F1 & Pr. & Re. & F1 & Pr. & Re. & F1 & Pr. & Re. & F1 & Pr. & Re. \\
    
    \midrule
    \multicolumn{13}{l}{\texttt{\textbf{Automatic Baseline Models}}} \\
    HeidelTime-auto & 0.53 & 0.63 & 0.46 & 0.41 & 0.56 & 0.32 & 0.49 & 0.66 & 0.39 & 0.15 & 0.60 & 0.09\\ 
    BiLSTM+CRF & 0.58 & 0.64 & 0.51 & 0.56 & 0.61 & 0.51 & 0.58 & 0.59 & 0.58 & 0.44 & 0.54 & 0.37\\
    mBERT & 0.56 & 0.61 & 0.51 & 0.56 & 0.62 & 0.51 & 0.60 & 0.56 & 0.64 & 0.59 & 0.64 & 0.55 \\ 
    XLMR-base & 0.64 & 0.69 & 0.59 & 0.51 & 0.58 & 0.46 & 0.59 & 0.59 & 0.59 & 0.43 & 0.60 & 0.34 \\ 
    XLMR-large & 0.69 & 0.70 & 0.68 & \textit{0.68} & \textit{0.71} & \textbf{0.66} & 0.71 & 0.69 & 0.73 & 0.66 & 0.70 & 0.63 \\ 
    \midrule
    \multicolumn{13}{l}{\texttt{\textbf{Projection Method}}} \\
    TMP-mBERT & 0.50 & 0.56 & 0.45 & 0.23 & 0.59 & 0.14 & 0.60 & 0.57 & 0.64 & / & / & / \\ 
    TMP-XLMRbase & 0.50 & 0.60 & 0.43 & 0.23 & 0.57 & 0.14 & 0.61 & 0.58 & 0.64 & / & / & / \\ 
    TMP-XLMRlarge & 0.52 & 0.61 & 0.46 & 0.24 & 0.59 & 0.15 & 0.61 & 0.58 & 0.63 & / & / & / \\ 
    \midrule
    \multicolumn{13}{l}{\texttt{\textbf{Transfer from EN (Ours)}}} \\
    XLTime-mBERT & 0.62 & 0.62 & 0.62 & 0.65 & 0.70 & 0.61 & 0.61 & 0.58 & 0.66 & 0.68 & 0.72 & 0.65 \\ 
    XLTime-XLMRbase & 0.67 & 0.67 & 0.68 & 0.60 & 0.63 & 0.58 & 0.64 & 0.62 & 0.66 & 0.64 & 0.68 & 0.60 \\
    XLTime-XLMRlarge & \textit{0.71} & \textbf{0.74} & 0.68 & \textbf{0.70} & \textbf{0.76} & \textit{0.65} & \textit{0.74} & \textit{0.71} & \textit{0.78} & \textit{0.72} & \textbf{0.79} & \textit{0.66} \\
    \midrule
    \multicolumn{13}{l}{\texttt{\textbf{Transfer from EN and others (Ours)}}} \\
    XLTime-mBERT & \textit{0.71} & 0.69 & 0.73 & \textit{0.68} & 0.69 & \textbf{0.66} & 0.73 & 0.70 & 0.76 & 0.68 & 0.72 & 0.65 \\ 
    XLTime-XLMRbase & 0.70 & 0.67 & \textit{0.74} & 0.65 & 0.69 & 0.62 & 0.66 & 0.64 & 0.68 & 0.70 & \textit{0.76} & 0.65 \\
    XLTime-XLMRlarge & \textbf{0.75} & \textit{0.72} & \textbf{0.78} & \textbf{0.70} & \textbf{0.76} & \textit{0.65} & \textbf{0.81} & \textbf{0.79} & \textbf{0.84} & \textbf{0.74} & \textbf{0.79} & \textbf{0.69} \\
    \midrule
    \multicolumn{5}{l}{\texttt{\textbf{Handcrafted Method}}}\\
    HeidelTime & 0.80 & 0.81 & 0.79 & 0.85 & 0.90 & 0.80 & 0.57 & 0.60 & 0.53 & / & / & /  \\
    \bottomrule
    \bottomrule
    \multirow{2}{*}{w/o type} &  \multicolumn{3}{c}{FR} & \multicolumn{3}{c}{ES} & \multicolumn{3}{c}{PT} & \multicolumn{3}{c}{EU}\\
    \cline{2-4} \cline{5-7} \cline{8-10} \cline{11-13}
    \multirow{2}{*}{} & F1 & Pr. & Re. & F1 & Pr. & Re. & F1 & Pr. & Re. & F1 & Pr. & Re. \\
    
    \midrule
    \multicolumn{13}{l}{\texttt{\textbf{Automatic Baseline Models}}} \\
    HeidelTime-auto & 0.55 & 0.65 & 0.47 & 0.42 & 0.58 & 0.33 & 0.50 & 0.67 & 0.39 & 0.17 & 0.66 & 0.10\\ 
    BiLSTM+CRF & 0.64 & 0.73 & 0.57 & 0.62 & 0.68 & 0.56 & 0.64 & 0.66 & 0.63 & 0.47 & 0.58 & 0.40\\
    mBERT & 0.63 & 0.70 & 0.58 & 0.62 & 0.69 & 0.56 & 0.66 & 0.63 & 0.69 & 0.65 & 0.71 & 0.60 \\ 
    XLMR-base & 0.69 & 0.75 & 0.64 & 0.54 & 0.61 & 0.48 & 0.63 & 0.64 & 0.62 & 0.46 & 0.64 & 0.36 \\ 
    XLMR-large & 0.75 & 0.78 & 0.73 & 0.72 & 0.75 & 0.69 & 0.75 & 0.74 & 0.76 & 0.70 & 0.74 & 0.67 \\ 
    \midrule
    \multicolumn{13}{l}{\texttt{\textbf{Projection Method}}} \\
    TMP-mBERT & 0.56 & 0.63 & 0.50 & 0.23 & 0.59 & 0.14 & 0.66 & 0.64 & 0.69 & / & / & / \\ 
    TMP-XLMRbase & 0.55 & 0.67 & 0.47 & 0.23 & 0.57 & 0.14 & 0.64 & 0.61 & 0.67 & / & / & / \\ 
    TMP-XLMRlarge & 0.56 & 0.66 & 0.50 & 0.24 & 0.59 & 0.15 & 0.65 & 0.61 & 0.68 & / & / & / \\ 
    \midrule
    \multicolumn{13}{l}{\texttt{\textbf{Transfer from EN (Ours)}}} \\
    XLTime-mBERT & 0.73 & 0.73 & 0.72 & 0.71 & 0.77 & 0.66 & 0.67 & 0.64 & 0.71 & 0.76 & 0.81 & 0.71 \\ 
    XLTime-XLMRbase & 0.78 & \textit{0.79} & 0.78 & 0.66 & 0.70 & 0.63 & 0.68 & 0.67 & 0.70 & 0.71 & 0.76 & 0.66 \\
    XLTime-XLMRlarge & 0.76 & \textit{0.79} & 0.73 & 0.72 & \textbf{0.79} & 0.67 & 0.77 & 0.74 & 0.81 & \textit{0.78} & \textit{0.85} & 0.71 \\
    \midrule
    \multicolumn{13}{l}{\texttt{\textbf{Transfer from EN and others (Ours)}}} \\
    XLTime-mBERT & 0.80 & 0.77 & \textit{0.82} & \textbf{0.77} & \textbf{0.79} & \textbf{0.74} & \textit{0.80} & \textit{0.77} & \textit{0.83} & 0.77 & 0.82 & 0.72 \\ 
    XLTime-XLMRbase & \textit{0.82} & \textit{0.79} & \textbf{0.86} & 0.72 & \textit{0.78} & 0.68 & 0.73 & 0.72 & 0.75 & \textbf{0.79} & \textbf{0.86} & \textit{0.73} \\
    XLTime-XLMRlarge & \textbf{0.84} & \textbf{0.82} & \textbf{0.86} & \textit{0.75} & \textbf{0.79} & \textit{0.71} & \textbf{0.84} & \textbf{0.82} & \textbf{0.87} & \textbf{0.79} & 0.84 & \textbf{0.74} \\
    \midrule
    \multicolumn{5}{l}{\texttt{\textbf{Handcrafted Method}}}\\
    HeidelTime & 0.86 & 0.87 & 0.85 & 0.86 & 0.91 & 0.81 & 0.60 & 0.64 & 0.57 & / & / & /  \\
    \bottomrule
  \end{tabular}
  \label{table:evaluation_low_resource_TEE_wt}
\end{table*}

\begin{table*}[t]
\caption{\small Low-resource language TEE with additional source languages (F1, precision, and recall scores w/ type). The \colorbox{blue!25}{blue cells} are expected to, while the \underline{underlined cells} actually outperform (by $\geq$ 3\%) using EN as the only source language. }
  \addtolength{\tabcolsep}{+3pt}
  \centering
  \small
    \begin{tabular}{l|c|ccc|c|ccc}
   \toprule
   \multicolumn{9}{c}{F1}\\
     \hline
     
     Target Language & \multicolumn{4}{c|}{FR} & \multicolumn{4}{c}{ES} \\
     \hline
     Source Language(s)& EN & EN, EU & EN, PT& EN, ES & EN & EN, EU & EN, PT & EN, FR \\
    \hline
     XLTime-mBERT & 0.62 & 0.61 & 0.61 & \cellcolor{blue!25}\underline{0.71} & 0.65 & 0.66 & 0.65 & \cellcolor{blue!25}\underline{0.68} \\
     XLTime-XLMRbase & 0.67 & 0.67 & 0.66 & \cellcolor{blue!25}\underline{0.70} & 0.60 & 0.61 & \underline{0.64} & \cellcolor{blue!25}\underline{0.65} \\
     XLTime-XLMRlarge & 0.71 & 0.73 & 0.73 & \cellcolor{blue!25}\underline{0.75} & 0.70 & 0.68 & 0.69 & \cellcolor{blue!25}0.68 \\
    \midrule
    Target Language & \multicolumn{4}{c|}{PT} & \multicolumn{4}{c}{EU} \\
     \hline
     Source Language(s)& EN & EN, FR & EN, ES& EN, EU & EN & EN, PT & EN, ES & EN, FR \\
    \hline
     XLTime-mBERT & 0.61 & \cellcolor{blue!25}\underline{0.72} & \cellcolor{blue!25}0.59 & \underline{0.73} & 0.68 & 0.66 & 0.66 & 0.68 \\
     XLTime-XLMRbase & 0.64 & \cellcolor{blue!25}0.66 & \cellcolor{blue!25}0.55 & 0.52 & 0.64 & 0.66 & 0.66 & \underline{0.70} \\
     XLTime-XLMRlarge & 0.74 & \cellcolor{blue!25}\underline{0.79} & \cellcolor{blue!25}\underline{0.81} & 0.71 & 0.72 & 0.71 & 0.74 & 0.72 \\

     \bottomrule
     \multicolumn{9}{c}{Precision}\\
     \hline
     Target Language & \multicolumn{4}{c|}{FR} & \multicolumn{4}{c}{ES} \\
     \hline
     Source Language(s)& EN & EN, EU & EN, PT& EN, ES & EN & EN, EU & EN, PT & EN, FR \\
    \hline
     XLTime-mBERT & 0.62 & 0.59 & 0.62 & \cellcolor{blue!25}\underline{0.69} & 0.70 & 0.69 & 0.71 & \cellcolor{blue!25}0.69 \\
     XLTime-XLMRbase & 0.67 & 0.66 & 0.67 & \cellcolor{blue!25}0.67 & 0.63 & 0.64 & \underline{0.67} & \cellcolor{blue!25}\underline{0.69} \\
     XLTime-XLMRlarge & 0.74 & 0.72 & 0.76 & \cellcolor{blue!25}0.72 & 0.76 & 0.65 & 0.73 & \cellcolor{blue!25}0.68 \\
    \midrule
    Target Language & \multicolumn{4}{c|}{PT} & \multicolumn{4}{c}{EU} \\
     \midrule
     Source Language(s)& EN & EN, FR & EN, ES& EN, EU & EN & EN, PT & EN, ES & EN, FR \\
    \hline
     XLTime-mBERT & 0.58 & \cellcolor{blue!25}\underline{0.68} & \cellcolor{blue!25}0.56 & \underline{0.70} & 0.72 & 0.70 & 0.69 & 0.72 \\
     XLTime-XLMRbase & 0.62 & \cellcolor{blue!25}0.64 & \cellcolor{blue!25}0.51 & 0.49 & 0.68 & \underline{0.73} & 0.69 & \underline{0.76} \\
     XLTime-XLMRlarge & 0.71 & \cellcolor{blue!25}\underline{0.75} & \cellcolor{blue!25}\underline{0.79} & 0.68 & 0.79 & 0.75 & 0.79 & 0.79 \\
    \bottomrule
     \multicolumn{9}{c}{Recall}\\
     \hline
     Target Language & \multicolumn{4}{c|}{FR} & \multicolumn{4}{c}{ES} \\
     \hline
     Source Language(s)& EN & EN, EU & EN, PT& EN, ES & EN & EN, EU & EN, PT & EN, FR \\
    \hline
     XLTime-mBERT & 0.62 & 0.62 & 0.59 & \cellcolor{blue!25}\underline{0.73} & 0.61 & 0.64 & 0.60 & \cellcolor{blue!25}\underline{0.66} \\
     XLTime-XLMRbase & 0.68 & 0.67 & 0.64 & \cellcolor{blue!25}\underline{0.74} & 0.58 & 0.59 & \underline{0.61} & \cellcolor{blue!25}\underline{0.62} \\
     XLTime-XLMRlarge & 0.68 & \underline{0.73} & \underline{0.71} & \cellcolor{blue!25}\underline{0.78} & 0.65 & \underline{0.71} & 0.65 & \cellcolor{blue!25}0.67 \\
    \midrule
    Target Language & \multicolumn{4}{c|}{PT} & \multicolumn{4}{c}{EU} \\
     \hline
     Source Language(s)& EN & EN, FR & EN, ES& EN, EU & EN & EN, PT & EN, ES & EN, FR \\
    \hline
     XLTime-mBERT & 0.66 & \cellcolor{blue!25}\underline{0.75} & \cellcolor{blue!25}0.62 & \underline{0.76} & 0.65 & 0.63 & 0.64 & 0.64 \\
     XLTime-XLMRbase & 0.66 & \cellcolor{blue!25}0.68 & \cellcolor{blue!25}0.60 & 0.55 & 0.60 & 0.60 & \underline{0.63} & \underline{0.65} \\
     XLTime-XLMRlarge & 0.78 & \cellcolor{blue!25}\underline{0.83} & \cellcolor{blue!25}\underline{0.84} & 0.74 & 0.66 & 0.67 & \underline{0.69} & 0.67 \\
    \bottomrule
  \end{tabular}
  \label{table:evaluation_source_languages_wt_f1_pr_rc}
\end{table*}

\begin{table*}[t]
\caption{\small Low-resource language TEE with additional source languages (precision and recall scores w/o type). The \colorbox{blue!25}{blue cells} are expected to, while the \underline{underlined cells} actually outperform (by $\geq$ 4\%) using EN as the only source language. }
  \addtolength{\tabcolsep}{+3pt}
  \centering
  \small
    \begin{tabular}{l|c|ccc|c|ccc}
   \toprule
     \multicolumn{9}{c}{Precision}\\
     \hline
     Target Language & \multicolumn{4}{c|}{FR} & \multicolumn{4}{c}{ES} \\
     \hline
     Source Language(s)& EN & EN, EU & EN, PT& EN, ES & EN & EN, EU & EN, PT & EN, FR \\
    \hline
     XLTime-mBERT & 0.73 & 0.76 & 0.76 & \cellcolor{blue!25}\underline{0.77} & 0.77 & 0.76 & 0.79 & \cellcolor{blue!25}0.79 \\
     XLTime-XLMRbase & 0.79 & 0.77 & 0.81 & \cellcolor{blue!25}0.79 & 0.70 & 0.72 & \underline{0.75} & \cellcolor{blue!25}\underline{0.78} \\
     XLTime-XLMRlarge & 0.79 & 0.81 & \underline{0.84} & \cellcolor{blue!25}0.82 & 0.79 & 0.70 & 0.79 & \cellcolor{blue!25}0.74 \\
    \midrule
    Target Language & \multicolumn{4}{c|}{PT} & \multicolumn{4}{c}{EU} \\
     \hline
     Source Language(s)& EN & EN, FR & EN, ES& EN, EU & EN & EN, PT & EN, ES & EN, FR \\
    \hline
     XLTime-mBERT & 0.64 & \cellcolor{blue!25}\underline{0.77} & \cellcolor{blue!25}0.67 & \underline{0.77} & 0.81 & 0.78 & 0.79 & 0.82 \\
     XLTime-XLMRbase & 0.67 & \cellcolor{blue!25}\underline{0.72} & \cellcolor{blue!25}0.60 & 0.54 & 0.76 & \underline{0.82} & 0.79 & \underline{0.86} \\
     XLTime-XLMRlarge & 0.74 & \cellcolor{blue!25}\underline{0.79} & \cellcolor{blue!25}\underline{0.82} & 0.72 & 0.85 & 0.85 & 0.84 & 0.84 \\
     
     \bottomrule
     \multicolumn{9}{c}{Recall}\\
     \hline
     Target Language & \multicolumn{4}{c|}{FR} & \multicolumn{4}{c}{ES} \\
     \hline
     Source Language(s)& EN & EN, EU & EN, PT& EN, ES & EN & EN, EU & EN, PT & EN, FR \\
    \hline
     XLTime-mBERT & 0.72 & \underline{0.77} & 0.69 & \cellcolor{blue!25}\underline{0.82} & 0.66 & 0.69 & 0.66 & \cellcolor{blue!25}\underline{0.74} \\
     XLTime-XLMRbase & 0.78 & 0.76 & 0.75 & \cellcolor{blue!25}\underline{0.86} & 0.63 & 0.64 & \underline{0.68} & \cellcolor{blue!25}\underline{0.68} \\
     XLTime-XLMRlarge & 0.73 & \underline{0.81} & \underline{0.77} & \cellcolor{blue!25}\underline{0.86} & 0.67 & \underline{0.75} & \underline{0.71} & \cellcolor{blue!25}\underline{0.72} \\
    \midrule
    Target Language & \multicolumn{4}{c|}{PT} & \multicolumn{4}{c}{EU} \\
     \hline
     Source Language(s)& EN & EN, FR & EN, ES& EN, EU & EN & EN, PT & EN, ES & EN, FR \\
    \hline
     XLTime-mBERT & 0.71 & \cellcolor{blue!25}\underline{0.83} & \cellcolor{blue!25}0.74 & \underline{0.83} & 0.71 & 0.69 & 0.70 & 0.72 \\
     XLTime-XLMRbase & 0.70 & \cellcolor{blue!25}\underline{0.75} & \cellcolor{blue!25}0.66 & 0.59 & 0.66 & 0.67 & \underline{0.70} & \underline{0.73} \\
     XLTime-XLMRlarge & 0.81 & \cellcolor{blue!25}\underline{0.87} & \cellcolor{blue!25}\underline{0.87} & 0.77 & 0.71 & 0.74 & 0.74 & 0.71 \\
    \bottomrule
  \end{tabular}
  \label{table:evaluation_source_languages_wot_pr_rc}
\end{table*}

\begin{table}[t]
\caption{\small Supervised English TEE on TE3 (w/ type | w/o type).}
  \centering
  \small
    \begin{tabular}{l|ccc}
   \toprule
     & F1 & Pr. & Re.\\
    
    \midrule
    \multicolumn{4}{l}{\texttt{\textbf{Rule-based Models}}} \\
    HeidelTime & 0.77| 0.81 &\textit{0.80}| 0.84  &0.75| 0.79 \\
    SynTime & 0.65| \textbf{0.92} &0.65| \textbf{0.91} & 0.66| \textbf{0.93} \\
    PTime & 0.67| \textit{0.85}&0.68| \textit{0.88} & 0.65| 0.83 \\
    \midrule
    \multicolumn{4}{l}{\texttt{\textbf{Language Models}}} \\
    BERT-base & 0.76| 0.82 & 0.78| 0.85 & 0.74| 0.80  \\
    BERT-large & \textbf{0.79}| 0.83 & 0.77| 0.82 & \textbf{0.80}| \textit{0.84} \\
    mBERT & \textbf{0.79}| 0.84 & \textit{0.80}| 0.86 & 0.77| 0.82  \\
    RoBERTa & \textit{0.78}| 0.84 & 0.79| 0.86 & 0.77| 0.82 \\
    XLMR-base & \textbf{0.79}| 0.81 & \textit{0.80}| 0.82 & 0.77| 0.81  \\
    XLMR-large & \textit{0.78}| 0.81 & 0.78| 0.82 & \textit{0.78}| 0.81 \\
    T5Encoder & \textbf{0.79}| 0.82 & \textbf{0.82}| 0.85 & \textit{0.78}| 0.80 \\
    \bottomrule
  \end{tabular}
  \label{table:evaluation_EN_TEE_TE3}
\end{table}

\begin{table}[t]
\caption{\small Supervised English TEE on Wikiwars (w/ type | w/o type).}
  \centering
  \small
    \begin{tabular}{l|ccc}
   \toprule
     & F1 & Pr. & Re.\\
    
    \midrule
    \multicolumn{4}{l}{\texttt{\textbf{Rule-based Models}}} \\
    HeidelTime & 0.80| 0.85  & 0.86| 0.92 & 0.75| 0.80  \\
    SynTime & 0.79| 0.79 &0.79| 0.79 & 0.79| 0.79  \\
    PTime & 0.86| 0.86 & 0.87| 0.87 & 0.86| 0.86 \\
    \midrule
    \multicolumn{4}{l}{\texttt{\textbf{Language Models}}} \\
    BERT-base & 0.94| 0.94 & \textit{0.95}| \textit{0.95} & 0.94| 0.94  \\
    BERT-large & 0.95| 0.95 & 0.94| 0.94 & 0.96| 0.96 \\
    mBERT & \textbf{0.97}| \textbf{0.97} & \textbf{0.96}| \textbf{0.96} & \textit{0.97}| \textit{0.97}\\
    RoBERTa & 0.95| 0.95 & 0.94| 0.94 & \textit{0.97}| \textit{0.97} \\
    XLMR-base & \textbf{0.97}| \textbf{0.97} & \textit{0.95}| \textit{0.95} & \textbf{0.98}| \textbf{0.98}  \\
    XLMR-large & \textit{0.96}| \textit{0.96} & 0.94| 0.94 & \textit{0.97}| \textit{0.97} \\
    T5Encoder &  \textit{0.96}| \textit{0.96} & \textit{0.95}| \textit{0.95} & \textit{0.97}| \textit{0.97} \\
    \bottomrule
  \end{tabular}
  \label{table:evaluation_EN_TEE_wikiwars}
\end{table}

\begin{table}[t]
\caption{\small Supervised English TEE on Tweets (w/ type | w/o type).}
  \centering
  \small
    \begin{tabular}{l|ccc}
   \toprule
     & F1 & Pr. & Re.\\
    
    \midrule
    \multicolumn{4}{l}{\texttt{\textbf{Rule-based Models}}} \\
    HeidelTime & 0.80| 0.80  & \textit{0.90}| 0.90  & 0.72| 0.72 \\
    SynTime & 0.63| 0.92 &0.62| 0.91 & 0.65| 0.95 \\
    PTime & 0.66| \textbf{0.95}& 0.65| \textbf{0.94} & 0.67| \textit{0.96}  \\
    \midrule
    \multicolumn{4}{l}{\texttt{\textbf{Language Models}}} \\
    BERT-base & \textit{0.92}| \textit{0.94} & \textit{0.90}| \textit{0.93} & 0.93| 0.95   \\
    BERT-large & 0.86| 0.92 & 0.84| 0.92 & 0.88| 0.92\\
    mBERT  & 0.87| 0.91 & 0.85| 0.88 & 0.90| 0.94 \\
    RoBERTa & 0.91| \textbf{0.95} & 0.89| \textit{0.93} & \textit{0.94}| \textbf{0.97}\\
    XLMR-base & 0.90| \textit{0.94}& 0.87| 0.92 & 0.93| \textbf{0.97} \\
    XLMR-large & \textbf{0.93}| \textbf{0.95} & \textbf{0.91}| \textit{0.93} & \textbf{0.95}| \textit{0.96} \\
    T5Encoder & 0.87| 0.93 & 0.84| 0.91 & 0.91| 0.95\\
    \bottomrule
  \end{tabular}
  \label{table:evaluation_EN_TEE_tweets}
\end{table}

\section{Language Models on English TEE}
\label{experiment:pretrained models}

In our early experiments, we reexamine the language models on English TEE. This section presents the results.
\subsection{Experimental Setup}
We study BERT \cite{devlin2018bert} and XLMR \cite{conneau2019unsupervised} variants, RoBERTa \cite{liu2019roberta} and T5 Encoder \cite{raffel2019exploring}. We compare them to rule-based methods including HeidelTime \cite{strotgen2013multilingual}, SynTime \cite{zhong2017time}, and PTime \cite{ding2019pattern}, which report SOTA performances on Wikiwars, TE3, and Tweets, respectively.
We experiment on both settings, i.e., ``with type" and ``without type", and report F1, precision, and recall in strict match \cite{uzzaman2013semeval}. We use the data splits following \citet{ding2019pattern} and the experimental settings introduced in Section~\ref{sec:experimental_setup}.

\subsection{Evaluation Results}
Tables~\ref{table:evaluation_EN_TEE_TE3}, \ref{table:evaluation_EN_TEE_wikiwars}, and \ref{table:evaluation_EN_TEE_tweets} show the results. We observe: \textbf{1)} When ignoring the types, the language models are inferior to SynTime on TE3, on par with or better than the rule-based methods on Wikiwars and Tweets. 
\textbf{2)} When considering the types, the language models outperform the previous SOTAs by 11-22\%, 18-21\%, and 30-41\% in F1 on TE3, Wikiwars, and Tweets datasets. 

\end{document}